\documentclass[letterpaper, 10 pt, conference]{ieeeconf}  

\IEEEoverridecommandlockouts                              

\usepackage{graphics} 
\usepackage{epsfig} 
\usepackage{mathptmx} 
\usepackage{times} 
\usepackage{amsmath} 
\usepackage{amssymb}  
\usepackage{subcaption}
\usepackage{cite}  
\usepackage[]{algorithm} 
\usepackage{algpseudocode} 
\usepackage{float}  
\usepackage{graphicx}
\usepackage{placeins}
\usepackage{balance}
\usepackage{xcolor}

\title{\LARGE \bf
Unsupervised Deep Persistent Monocular Visual Odometry and Depth Estimation in Extreme Environments
}

\author{Yasin Almalioglu$^{1,2}$, Angel Santamaria-Navarro$^{2}$, Benjamin Morrell$^{2}$, and Ali-akbar Agha-mohammadi$^{2}$
\thanks{$^{1}$Y. Almalioglu is with the Computer Science Department, The University of Oxford, UK. {\tt\small \{yasin.almalioglu\}@cs.ox.ac.uk}}%
\thanks{$^{2}$Y. Almalioglu, A. Santamaria-Navarro, B. Morrell and A. Agha-mohammadi are with the NASA - Jet Propulsion Laboratory, California Institute of Technology, Pasadena, CA, US. {\tt\small \{yasin.almalioglu, angel.santamaria.navarro, benjamin.morrell, aliakbar.aghamohammadi\}@jpl.nasa.gov}}
\thanks{This work was carried out at the Jet Propulsion Laboratory, California Institute of Technology, under a contract with the National Aeronautics and Space Administration.
Copyright 2020 California Institute of Technology. U.S. Government sponsorship acknowledged.}%
}

\begin{document}

\maketitle
\thispagestyle{empty}
\pagestyle{empty}

\begin{abstract}
In recent years, unsupervised deep learning approaches have received significant attention to estimate the depth and visual odometry (VO) from unlabelled monocular image sequences. However, their performance is limited in challenging environments due to perceptual degradation, occlusions and rapid motions. Moreover, the existing unsupervised methods suffer from the lack of scale-consistency constraints across frames, which causes that the VO estimators fail to provide persistent trajectories over long sequences. In this study, we propose an unsupervised monocular deep VO framework that predicts six-degrees-of-freedom pose camera motion and depth map of the scene from unlabelled RGB image sequences. 
We provide detailed quantitative and qualitative evaluations of the proposed framework on a) a challenging dataset collected during the DARPA Subterranean challenge\footnote{https://www.subtchallenge.com}; and b) the benchmark KITTI and Cityscapes datasets.
The proposed approach outperforms both traditional and state-of-the-art unsupervised deep VO methods providing better results for both pose estimation and depth recovery.
The presented approach is part of the solution used by the COSTAR team participating at the DARPA Subterranean Challenge.
\end{abstract}

\section{INTRODUCTION}

Autonomous robot traversal and 3D structure reconstruction capabilities have a wide variety of applications in extreme environments, such as autonomous driving~\cite{geiger2012we}; search and rescue in emergency responses~\cite{nagatani2013emergency}; inspection of underground habitats~\cite{santamarianavarro2020resilient, ebadi2020lamp}; or planetary surface exploration~\cite{agha2019robotic, sasaki2020map}.
The ability to estimate ego-motion and the scene map is critical to enable these capabilities. 
In this sense, vision-based solutions for localization and 3D structure reconstruction are prevailing thanks to the camera characteristics, being low cost; with low weight and low power consumption; and offering reasonably rich exteroceptive information.

Camera motion estimation and depth map reconstruction are fundamental and well-studied problems in computer vision.
Many traditional techniques have been proposed in the last decade, achieving reasonably good results~\cite{newcombe2011dtam, klein2007parallel, furukawa2010towards, turan2018sparse, almalioglu2020milli}.
However, they are usually committed to finding accurate image correspondences between consecutive frames, which is a frequently violated condition in challenging environments.
For instance, such matching can only be established for a subset of all pixels, which leaves the problem of estimating ill-posed depth.
These scenarios typically involve off-nominal conditions such as perceptual degradation; variable lighting conditions; non-Lambertian surfaces or variable surface colors and textures; potential presence of obscurants (e.g., fog, smoke, dust or water puddles); and physical obstructions within the field-of-view~\cite{davide2011visual, lowe2004distinctive}.


Following the success of deep learning in different domains, recent approaches solve the ill-posed depth estimation by using data-driven techniques.
Even if the data is insufficient to resolve ambiguities, deep networks
can estimate the camera pose and depth maps by generalizing from prior examples they have learned~\cite{wang2017deepvo, dosovitskiy2015flownet}.
In this sense, supervised deep-learning-based methods have shown good performance, successfully alleviating issues such as scale drift, which affects traditional feature extraction and parameter tuning~\cite{clark2017vinet, saputra2020deeptio, turan2018deependovo, saputra2019learning}. 
Eigen et al. \cite{eigen2014depth} show that a Convolution Neural
Network (CNN) can predict the depth map from a single image using the ground truth depths acquired by range sensors. 
Although the supervised approaches \cite{eigen2014depth, kuznietsov2017semi, liu2015learning} show high-quality motion and depth estimation results, the acquisition of ground truth can be either impractical or even impossible in real-world scenes.

In recent years, unsupervised deep learning approaches have achieved remarkable results, comparable to those from supervised techniques~\cite{zhou2017unsupervised, godard2017unsupervised, ummenhofer2017demon, turan2018unsupervised, mahjourian2018unsupervised, almalioglu2019selfvio, almalioglu2019ganvo}.
Unsupervised approaches allow learning from raw camera frames alone, without the need for supervision signals (e.g., depth sensors) and the trained networks are able to infer a depth map from a single image and ego-motion from consecutive images.
SfM-Learner \cite{zhou2017unsupervised} is among the first unsupervised methods that jointly learn camera motion and depth estimation.
Geonet \cite{yin2018geonet} and Ranjan et al. \cite{ranjan2019competitive} incorporate optical flow into the joint unsupervised training framework. SC-SFM \cite{bian2019unsupervised} enforces depth consistency to solve the scale
inconsistency issue in SfM-Learner \cite{zhou2017unsupervised}.

Although existing unsupervised learning methods provide state-of-the-art performance, their estimations are still limited in challenging environments.
Some visual degradation might violate their underlying frame correspondence assumptions that use geometric image reconstruction.
Further, and more importantly, recent works suffer from the per-frame scale ambiguity due to the lack of a single and consistent scaling of the camera motion. As a result, the ego-motion network cannot predict a full camera trajectory over a long image sequence.
Multiple approaches propose to disconnect the geometric constraints from the unsupervised architecture to handle occlusions in optical flow estimation \cite{wang2018occlusion, janai2018unsupervised}. On the other hand, differentiable mesh rendering \cite{nguyen2018rendernet, kato2018neural} offers an alternative geometric approach to handle occlusions. In the context of joint learning of depth recovery and ego-motion estimation, several works propose a learned explainability mask \cite{zhou2017unsupervised} by penalizing the minimum re-projection loss between the frames or use optical flow \cite{yang2018every} to solve occlusion problems.
Gordon et al. \cite{gordon2019depth} propose a geometric method for occlusion handling.
Drawing inspiration of some of these methods, we address both the occlusion problem and scale ambiguity across frames without incurring a substantial additional computational cost.


In this study, we propose a novel monocular visual odometry estimation and depth recovery approach that can operate in challenging environments, able to produce persistent results over a long duration.
We train an unsupervised deep neural network that 
takes a sequence of monocular images and estimates 6-Degrees-of-Freedom (DoF) camera motion and the depth map.
Similar to~\cite{szeliski1999prediction, zhan2018unsupervised, almalioglu2019ganvo}, we utilize for the training a view reconstruction approach as part of the objective function.
The entire pose estimation and depth map reconstruction pipeline is a persistent framework thanks to the occlusion-aware and scale-aware objectives imposed during the optimization of the network.

In summary, the main contributions of our method are the following: 
\begin{itemize}
    \item Two new loss functions to tackle the problems of occlusions and trajectory scale. Further, we describe the total loss function to incorporate them into the unsupervised architecture. These contributions alleviate the need for separate networks to handle occlusions and scale-ambiguity across frames.
    \item A novel depth enhancement technique for unsupervised depth reconstruction methods, which enable the generation of depth images in challenging environments.
\end{itemize}
These contributions enable long-duration operations in perceptually degraded environments, which to the best of authors' knowledge, is the first unsupervised deep-learning approach to estimate the camera (robot) odometry while reconstructing the depth map using images from a monocular camera.

To validate the proposed approach, we evaluate it on the KITTI~\cite{geiger2013vision} and Cityscapes~\cite{cordts2016cityscapes} datasets as benchmarks for comparative analysis with other state-of-the-art methods.
This evaluation criterion has been widely accepted in the robotics community in recent years.
This approach is part of the state estimation framework developed by the team CoSTAR\footnote{https://costar.jpl.nasa.gov} for the DARPA Subterranean Challenge\footnote{https://subtchallenge.com}.
Hence, we also show results on a dataset from the NASA-JPL, California Institute of Technology, with images captured by a Husky Clearpath\footnote{https://clearpathrobotics.com} robot under perception-challenging conditions, during the exploration of the underground urban circuit of the DARPA Challenge.

The rest of this paper is organized as follows.
Section \ref{sec:method} presents the proposed approach, with detailed descriptions of the architecture and its mathematical background.
Section \ref{sec:results} shows our quantitative and qualitative results with comparisons to the existing methods in the literature.
Finally, Section \ref{sec:conclusion} briefly discusses the findings and concludes the study.
\section{Unsupervised Depth and Pose Estimation}
\label{sec:method}

\begin{figure*}
\includegraphics[width=\textwidth]{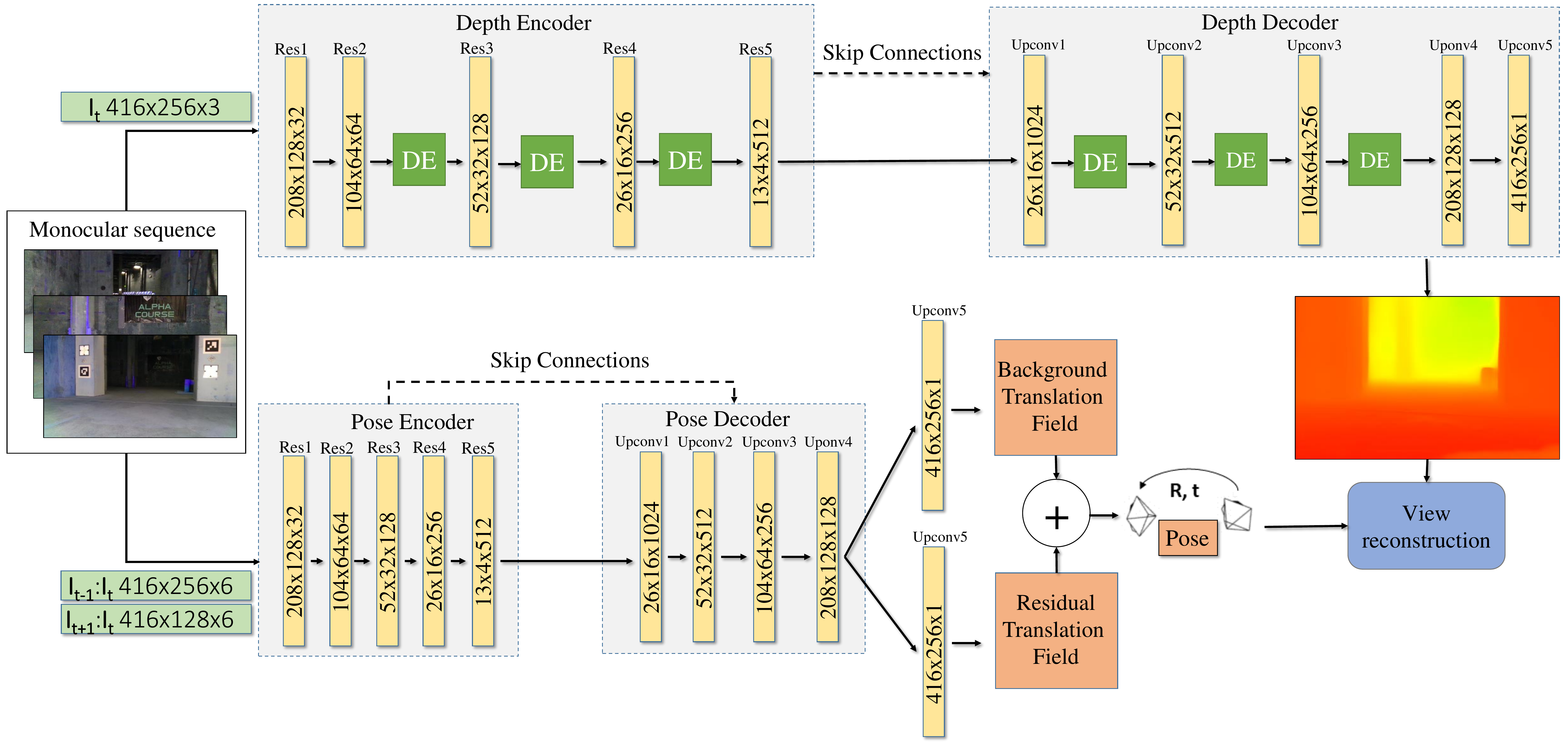}
\caption{Proposed unsupervised deep-learning architecture for pose estimation and depth map generation.
The spatial dimensions on layers and output channels show the tensor shapes that flow through the network.
The depth network generates a depth map from a single input image, using our depth enhancement module (DE).
The pose network estimates a background and a residual motion field of the given three consecutive frames. 
The network is optimized using scale-aware and occlusion-aware loss functions along with photometric and smoothness losses.}
\label{fig:deeparchitecture}
\end{figure*}

\subsection{Architecture Overview}
\label{sec:overview}
The proposed architecture is based on unsupervised deep learning to learn ego-motion and depth from monocular image sequences jointly.
The raw RGB sequences, consisting of a target and source views, are stacked together to form an input batch to the multi-view pose estimation and depth recovery modules. 
The motion-prediction network predicts a motion of every pixel with respect to the background and a residual translation field to account for moving objects.
In parallel, a second network generates a depth map of the target view.
The view reconstruction module reconstructs the target image using the predicted depth map, estimated 6-DoF camera pose and nearby colour values from source images. In this architecture, a) we impose scale-consistency across consecutive frames through a geometry consistent loss function;
b) we estimate occlusions geometrically, based on the estimated depth maps to apply this loss only in non-occluded pixels;
c) we regularize motion fields based on residual translations that indicate which pixels might belong to moving objects;
and d) we include other state-of-the-art loss functions to handle dissimilarity or edge-aware smoothness in a total loss function.
Furthermore, e) we introduce spatial–channel combinational attention into geometry understanding to explore the effectiveness of self-attention for scene geometry understanding.
This architecture is shown in Fig. \ref{fig:deeparchitecture} and its details are explained hereafter.

\begin{figure}[t]
\centering
\includegraphics[width=\columnwidth]{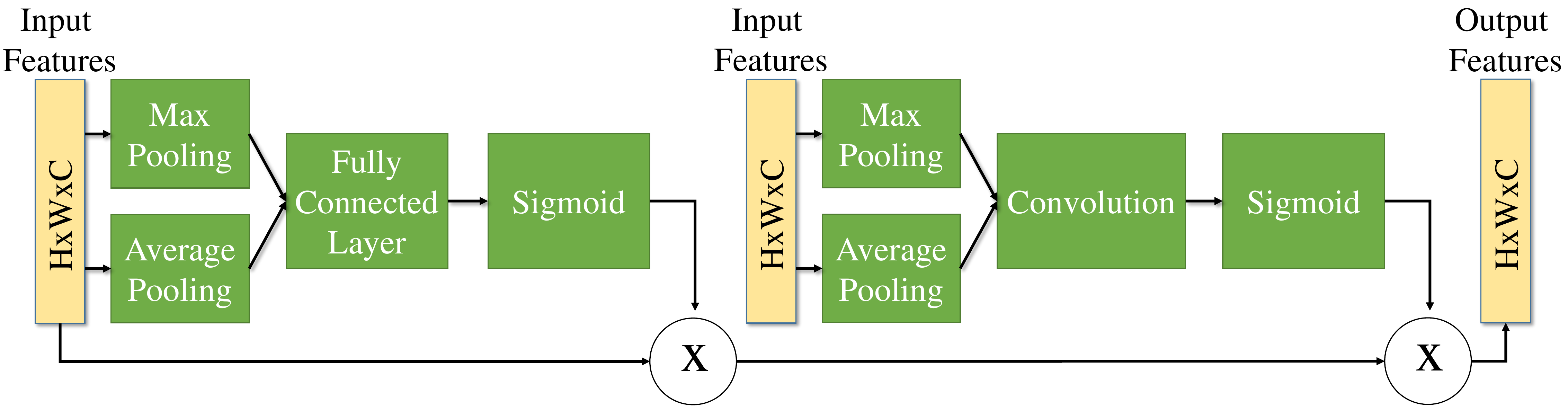}
\caption{Architecture of the depth enhancement module. } 
\label{fig:de_module}
\end{figure}

\subsection{Networks}
We rely on two convolutional networks based on the \mbox{ResNet-18} model~\cite{he2016deep}, one predicting depth from a single image, and the other predicting ego-motion and the motion field relative to the scene, using three input images.

\paragraph{Depth Network}
The first part of the architecture is a depth network that recovers a single-view depth map of the target frame. 
The depth prediction network uses a UNet architecture and a softplus activation ($z=\log(1+\mathrm e^\ell)$) to convert the logits ($\ell$) to depth values ($z$).
We embed depth enhancement modules (DE) into both encoder and decoder sub-networks, which re-calibrate depth features and can produce more useful and important features to capture fine details in the scene.

\textbf{Depth Enhancement Module:} Given the original feature map $F = \{F_{1}, F_{2},\ldots ,F_{c}\}$, where $c$ is the number of channels, the DE module produces a channel attention map $A_{c}$ and a spatial attention map $A_{s}$ to refine $F$ as shown in Fig. \ref{fig:de_module}. The max-pooling and average-pooling operations aggregate the global information of input features. Then, we feed these two features $F^{max}$ and $F^{avg}$ into a fully-connected layer with one hidden layer to recover the original channel size. 

\paragraph{Pose Network}
The second network shown in the bottom of Fig. \ref{fig:deeparchitecture} tries to estimate relative pose $\textbf{p} \in\mathbb{SE}(3)$ introduced by motion fields across frames. 
The motion estimation network is a UNet architecture based on FlowNet \cite{dosovitskiy2015flownet}. A stack of pose encoder and decoder sub-networks predicts the global rotation angles ($r_0$) and the global translation vector ($t_0$) that represent the movement of the entire scene with respect to the camera due to ego-motion. The decoder layers progressively refine the translation, from a single vector to a residual translation vector field $\delta t(x, y)$.
The translation field is defined as the sum of the global translation vector plus the masked residual translation:
\begin{equation}
t(x, y) = t_0 + m(x, y)\delta t(x, y),
\end{equation}
where $m(x, y)$ equals one at pixels that could belong to mobile objects and zero otherwise as described in Sec.~\ref{sec:occlusion-aware}.

\subsection{Loss functions}
\label{sec:occlusion-aware}
\paragraph{Occlusion-aware loss}
When the camera and the scene move relatively to each other, points in the scene that are visible in one frame may become occluded in another. The cross-frame consistency cannot be enforced on the occluded pixels by a loss. Given a depth map and a motion field in one frame, we geometrically detect where occlusions occur, and exclude the occluded areas from the consistency loss. The occlusion-aware loss re-projects the depth values onto the camera frame and detects if the depth value on the re-projected is visible. Gordon et al. \cite{gordon2019depth} propose to asymmetrically choose source points that land in front of the depth map in the target frame. However, the projected points at the occluded areas need interpolation that can fall into a region instead of specific locations. We propose to choose points that fall within the neighborhood distance $d_n$ of the occluded area. We also choose points that not only fall in front of the target map but also behind it to obtain a symmetric mask, which eliminates unnecessary reversed source-target depth computation. 

\paragraph{Scale-aware loss}
Given source and target depth maps $D_a$ and $D_b$, and the relative pose $P_{ab}$, we minimize the difference between the re-projected 3D scene structure:
\begin{equation}\label{eqn-depthdiff}
D_{\text{diff}}(p) = \frac{|D_b^{a}(p) - D'_b(p)|}{D_b^{a}(p) + D'_b(p)}
\end{equation}
where $D_b^{a}$ is the computed depth map of $I_b$ by warping $D_a$ using $P_{ab}$; and $D'_b$ is the re-projected depth map from $D_b$.
This optimization imposes a scale consistency constraint in the entire sequence as previously shown by Bian et al. \cite{bian2019unsupervised} using a point-wise distance across all pixels. Unlike \cite{bian2019unsupervised} that uses an occlusion mask based on the depth difference, we geometrically handle the occluded areas as explained in Sec.~\ref{sec:occlusion-aware}, which is more sensitive to fine details in the depth map (see Fig. \ref{fig:res_depth_kitti} for example results).
With the scale-aware training of the network, the pose network predicts globally scale-consistent trajectories even in challenging environments.

\paragraph{Total loss}
Previous works~\cite{zhou2017unsupervised, yin2018geonet, zou2018df, ranjan2019competitive} leveraged the brightness constancy and spatial smoothness priors proposed in~\cite{baker2004lucas}, and have showed how the photometric error between the warped and the target frames is effective in an unsupervised loss function to optimize the network.
We apply an occlusion-aware L1 loss for the photometric error due to its robustness to outliers.
In addition, we impose occlusion-aware cycle consistency for the predicted motion fields. We require that the inverse motion field is the opposite of the inverse motion. 
We add an additional image dissimilarity loss SSIM~\cite{wang2004image} to handle the varying ambient lighting in a complex environment as it normalizes the pixel illumination.
Finally, we include the edge-aware smoothness loss used in~\cite{ranjan2019competitive} to compensate for the inferior performance of the photometric loss in low-texture and non-homogeneous regions:
\begin{equation}
L_{s} = \sum_{p} ( e^{-\nabla I_a(p)} \cdot \nabla D_a(p) ) ^2,
\end{equation}
where $\nabla$ is the first derivative along spatial directions, which guides the smoothness by the edge of images.

\section{Experimental Results}
\label{sec:results}

We implemented our unsupervised architecture with the publicly available Tensorflow framework \cite{abadi2016tensorflow}.
We optimized the weights of the network using Adam optimization with the parameters $\beta_1 = 0.9$, $\beta_2 = 0.999$, learning rate of $0.001$ and mini-batch size of $8$.
We used sequential images of size $416 \times 256$ as the input tensors of the model.
We trained the model on an NVIDIA TITAN V model GPU.

\begin{table}[t]
\centering
    \begin{tabular}{l | c c | c c}
     \hline
     Methods & \multicolumn{2}{c|}{Seq. 09} & \multicolumn{2}{c}{Seq. 10} \\
     & $t_{err}$ ($\%$) & $r_{err}$ ($^{\circ}/100m$) & $t_{err}$ ($\%$) & $r_{err}$ ($^{\circ}/100m$) \\
     \hline
    ORB-SLAM~\cite{mur2015orb} & 15.30 & 0.26 & 3.68 & 0.48 \\
    Zhou~et al.~\cite{zhou2017unsupervised} & 17.84 & 6.78 & 37.91 & 17.78 \\
    Zhan~et al.~\cite{zhan2018unsupervised} & 11.93 & 3.91 & 12.45 & \textbf{3.46} \\
    GANVO \cite{almalioglu2019ganvo} & 11.52 & 3.53 & 11.60 & 5.17 \\
    SC-SFM \cite{bian2019unsupervised}  & 11.2  & 3.35 & 10.1 & 4.96 \\
    Ours  & \textbf{10.87}  & \textbf{3.14} & \textbf{8.91} & 4.45 \\
     \hline
    \end{tabular}
    \caption{Visual odometry results on KITTI~\cite{geiger2013vision} odometry dataset.
    We report the performance of ORB-SLAM~\cite{mur2015orb} as a reference to compare with state-of-the-art deep learning methods.
    }
    \label{tab:vo}
\end{table}

\begin{figure}[t]
	\begin{subfigure}{\columnwidth} 
	    \includegraphics[width=0.9\columnwidth]{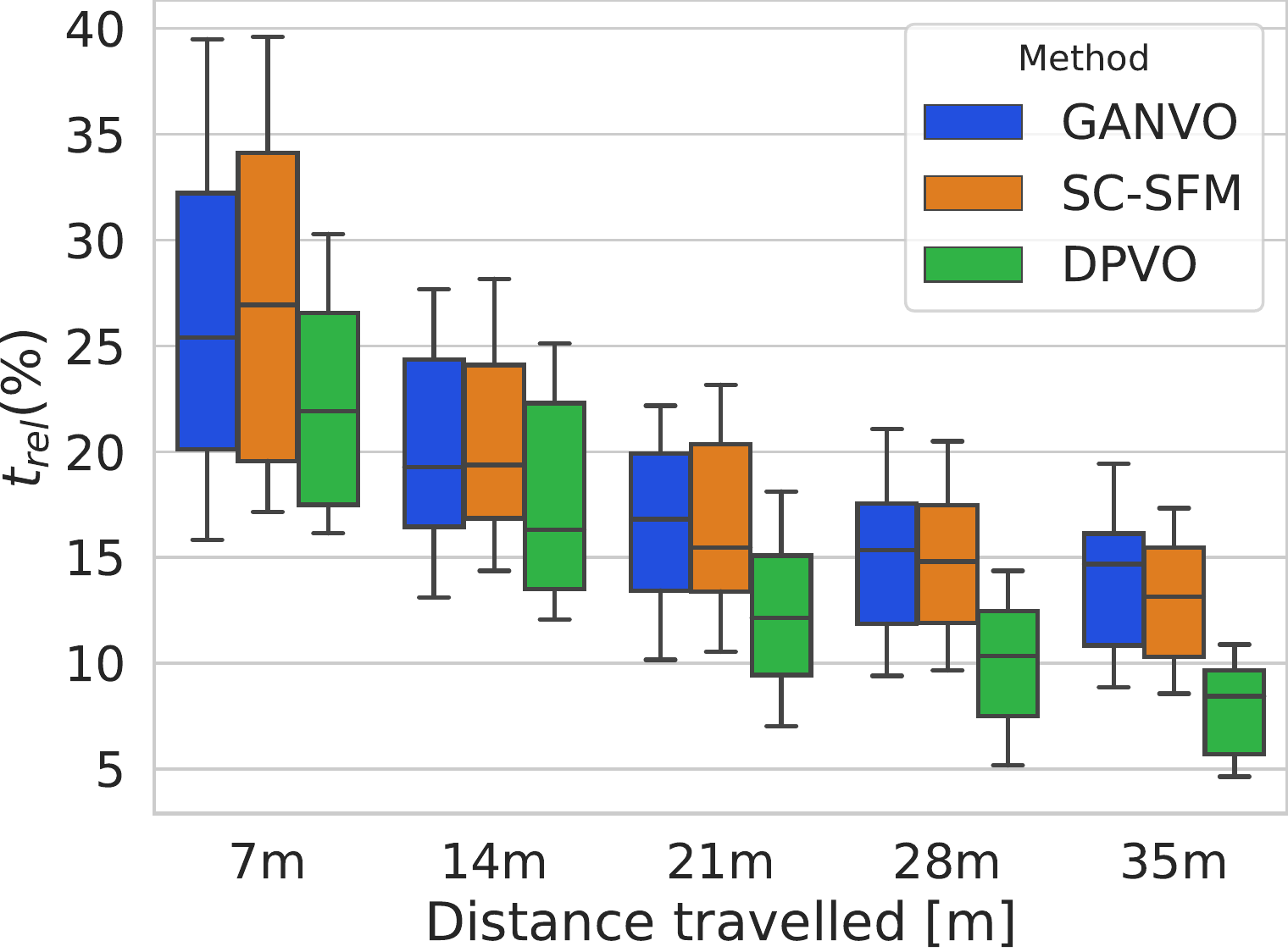}
        \label{fig:seq09}
	\end{subfigure}
	~
	\begin{subfigure}{\columnwidth} 
	    \includegraphics[width=0.9\columnwidth]{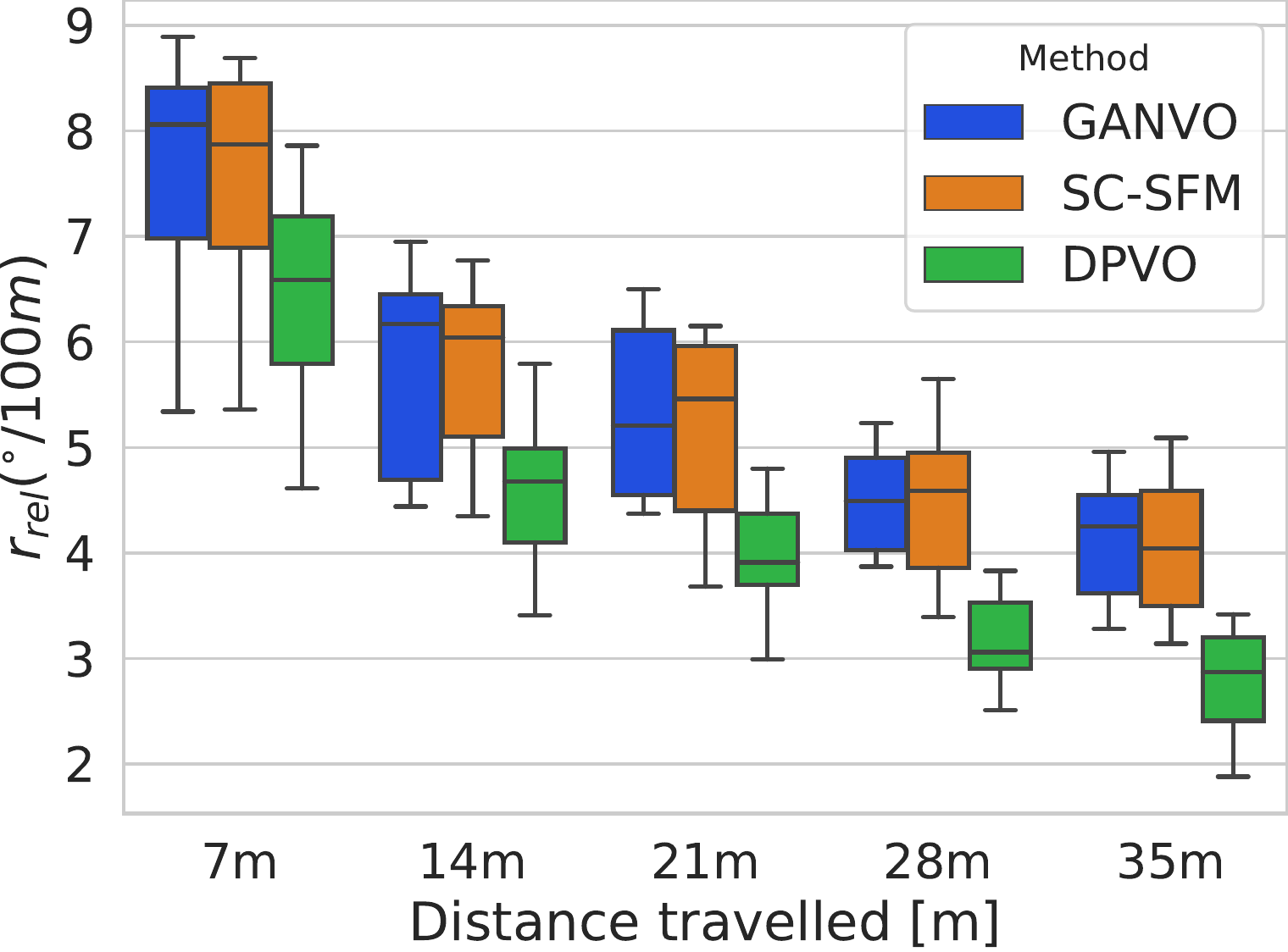}
        \label{fig:seq10}
	\end{subfigure}	
	\caption{Relative pose errors for the subterranean dataset, using different segment lengths ($\{7, 14, 21, 28, 35\}$ m) based on the shortest sequence throughout the whole trajectory in multiple segments (in total: 1h run and ~2km traversed), where our proposed approach (DPVO) outperforms existing deep-learning state-of-art methods.}
    \label{fig:rpe}
\end{figure}

The validation of the proposed approach is two fold. 
On the one hand we use the KITTI~\cite{geiger2012we} and the Cityscapes~\cite{cordts2016cityscapes} datasets for benchmarking, where we compare our method with standard training/test splits for the odometry and monocular depth map estimation tasks.
Second, we evaluate our method on a perception-challenging dataset recorded during the DARPA Subterranean Challenge, in order to prove the effectiveness of our architecture for both depth map reconstruction and pose estimation over long sequences in complex environments.
This subterranean dataset was gathered during the DARPA competition during an autonomous exploration of an underground environment in the Satsop Nuclear Plant, Elma, Washington.

\subsection{Pose estimation benchmark}
We evaluated the ego-motion prediction performance on the standard KITTI visual odometry split. Specifically, the KITTI sequences 09-10.
In this sense, the standard 5-point Absolute Trajectory Error (ATE) metric~\cite{zhou2017unsupervised, casser2019unsupervised, yin2018geonet} measures local agreement between the estimated trajectories and the respective ground truth. However, we believe that in this case, a relative pose error metric is better suited to measure the drift of an odometry system \cite{qin2018vins, leutenegger2015keyframe, gordon2019depth, zhan2018unsupervised}.
Thus, we show statistics for the relative translation and rotation error, divided by the distance travelled and averaged over the trajectory segments of lengths $\{7, 14, 21, 28, 35\}$ m over all sequences based on the shortest sequence.
Table \ref{tab:vo} summarizes both metrics.
As shown in Table \ref{tab:vo}, the proposed method outperforms all the competing unsupervised baselines on the KITTI sequences 09-10, without any need for global optimization steps such as loop closure detection, bundle adjustment and re-localization, revealing that out method persistently predicts ego-motion over long sequences. 
Since most of the compared methods are monocular approaches and lack a scaling factor to match with real-world scale, we scaled and aligned (7DoF optimization) the predictions to the ground truth associated poses during the evaluation by minimizing ATE~\cite{umeyama1991least}.

\begin{figure}[t]
 \vspace{-2mm}
 \centering
	\begin{subfigure}{0.9\columnwidth} 
	    \includegraphics[width=1.0\columnwidth]{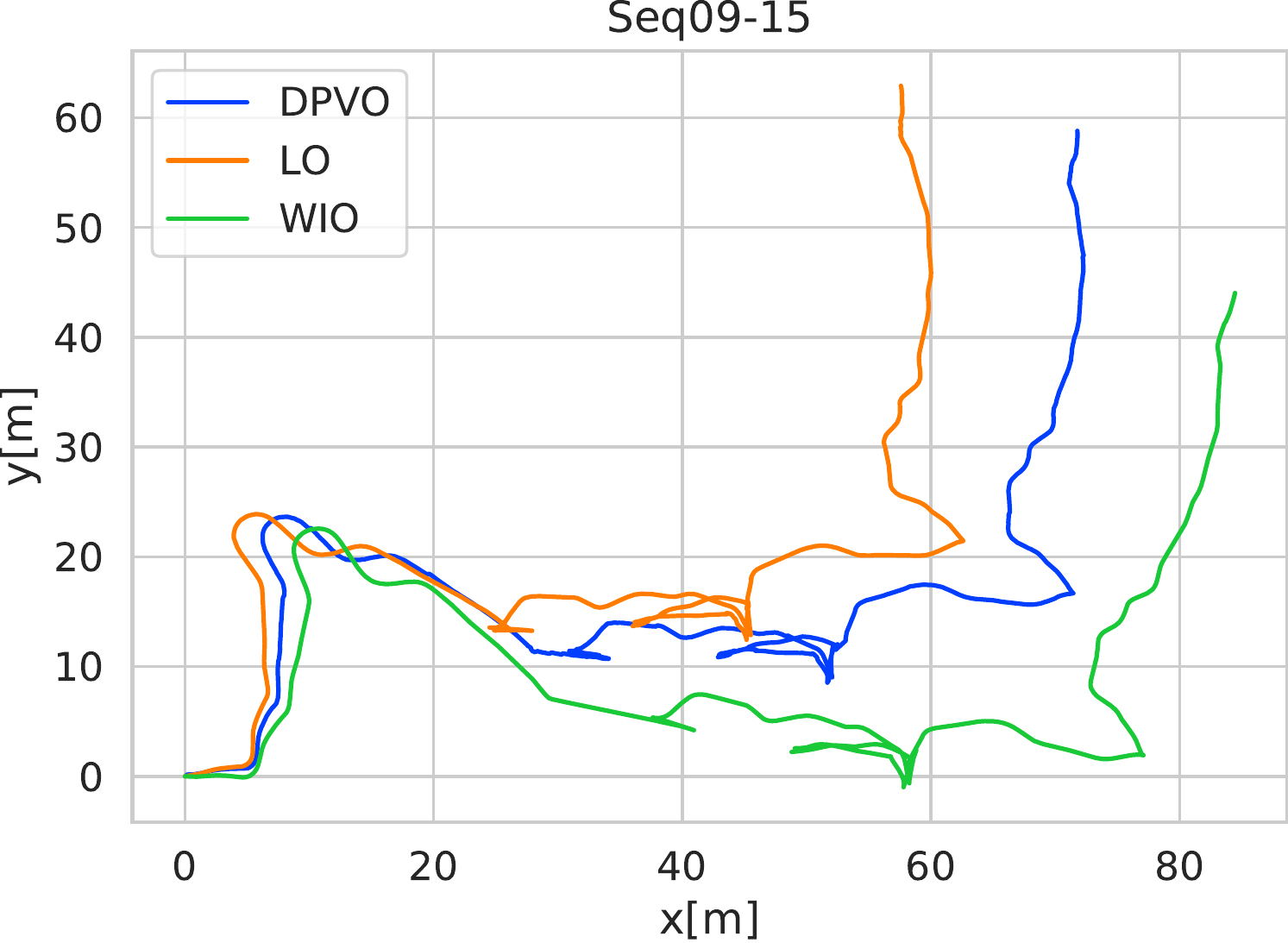}
        \caption{} 
        \label{fig:seq09-15}
	\end{subfigure}
	
	\begin{subfigure}{0.9\columnwidth} 
	    \includegraphics[width=1.0\columnwidth]{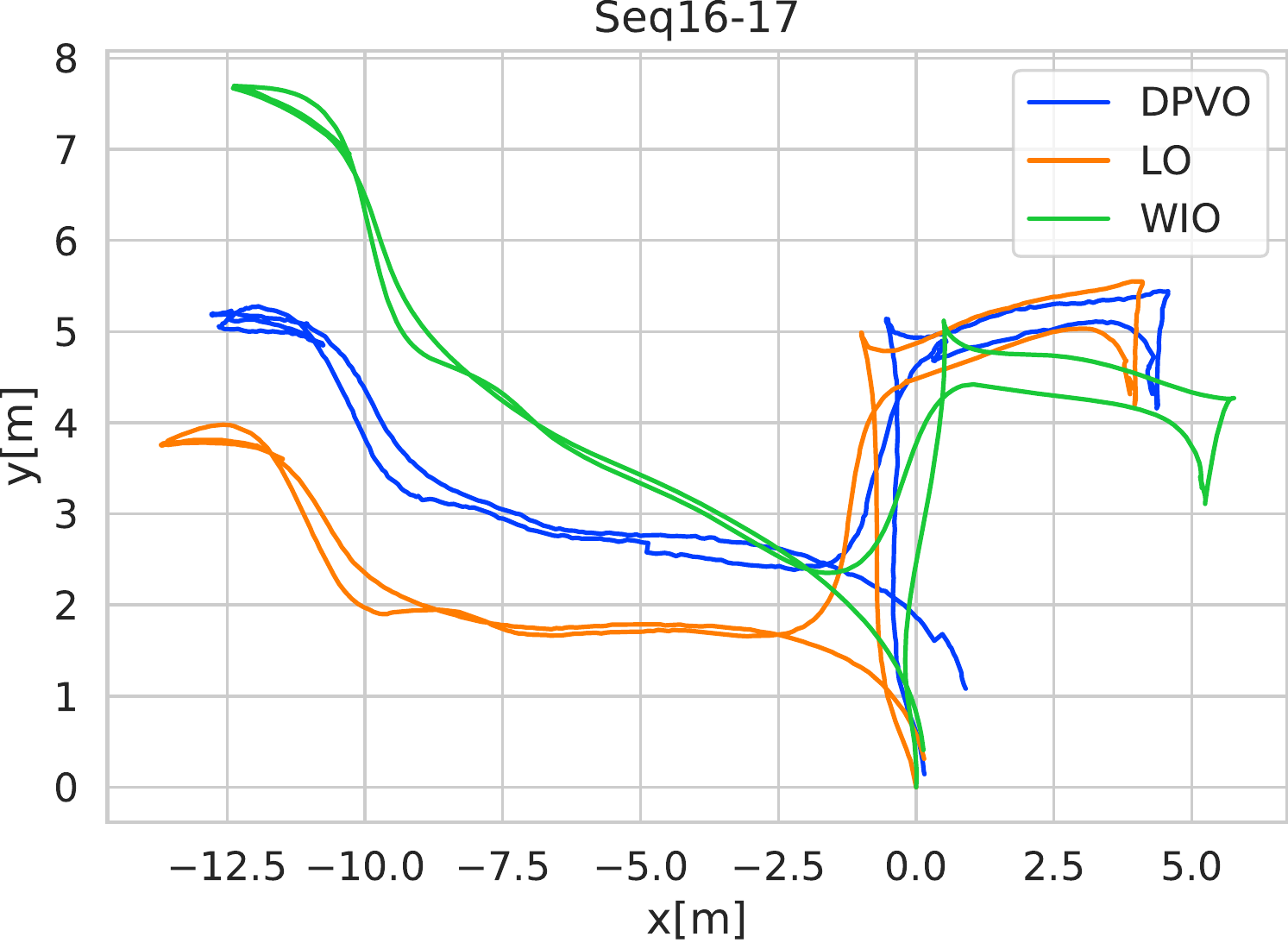}
        \caption{} 
        \label{fig:seq16-17}
	\end{subfigure}	
	\caption{Two sample trajectories from the statistical analysis using the dataset recorded by the COSTAR team during the DARPA Subterranean Challenge. Here, we compare our unsupervised learning method (DPVO) with a very accurate LIDAR odometry~\cite{Palieri} and a less-accurate wheel-inertial odometry, as ground-truth estimates are not available for this environment. DPVO is more resistant to drifts than wheel-inertial odometry, achieving performances comparable to those from the LIDAR odometry, in both rotational and translational motions.}
    \label{fig:seqs}
\end{figure}

Furthermore, we evaluated our approach on our challenging subterranean dataset (DARPA Challenge) the standard analysis criteria to show how it persistently estimates the ego-motion over a long duration in complex environments.
Fig. \ref{fig:rpe} shows the results of analyzing sub-sequences of lengths $\{7, 14, 21, 28, 35\}$ m and reports the average translational error $t_{err} (\%)$ and rotational errors $r_{err} (^\circ/100m)$.
As seen in Fig. \ref{fig:rpe}, our approach outperforms state-of-the-art methods in terms of both average translational error $t_{err} (\%)$ and rotational errors $r_{err} (^\circ/100m)$.
Moreover, to validate the usefulness of the approach we present in Fig.~\ref{fig:seqs} two sample sequences (two evaluated segments) from this subterranean dataset.
In this case, we compare our approach (DPVO) with a highly accurate LIDAR odometry ($360^o$ field-of-view)~\cite{Palieri} and (less-accurate) wheel-inertial odometry as baselines.
The sequence in Fig. \ref{fig:seq09-15} has $208.14$ m length, which shows the odometry estimation performance of DPVO over long subterranean sequences.
The sequence in Fig. \ref{fig:seq16-17} has $54.40$ m length and contains complex camera motions, proving that DPVO is resistant to abrupt motions.

\begin{figure}[t]
\centering
\includegraphics[width=\linewidth]{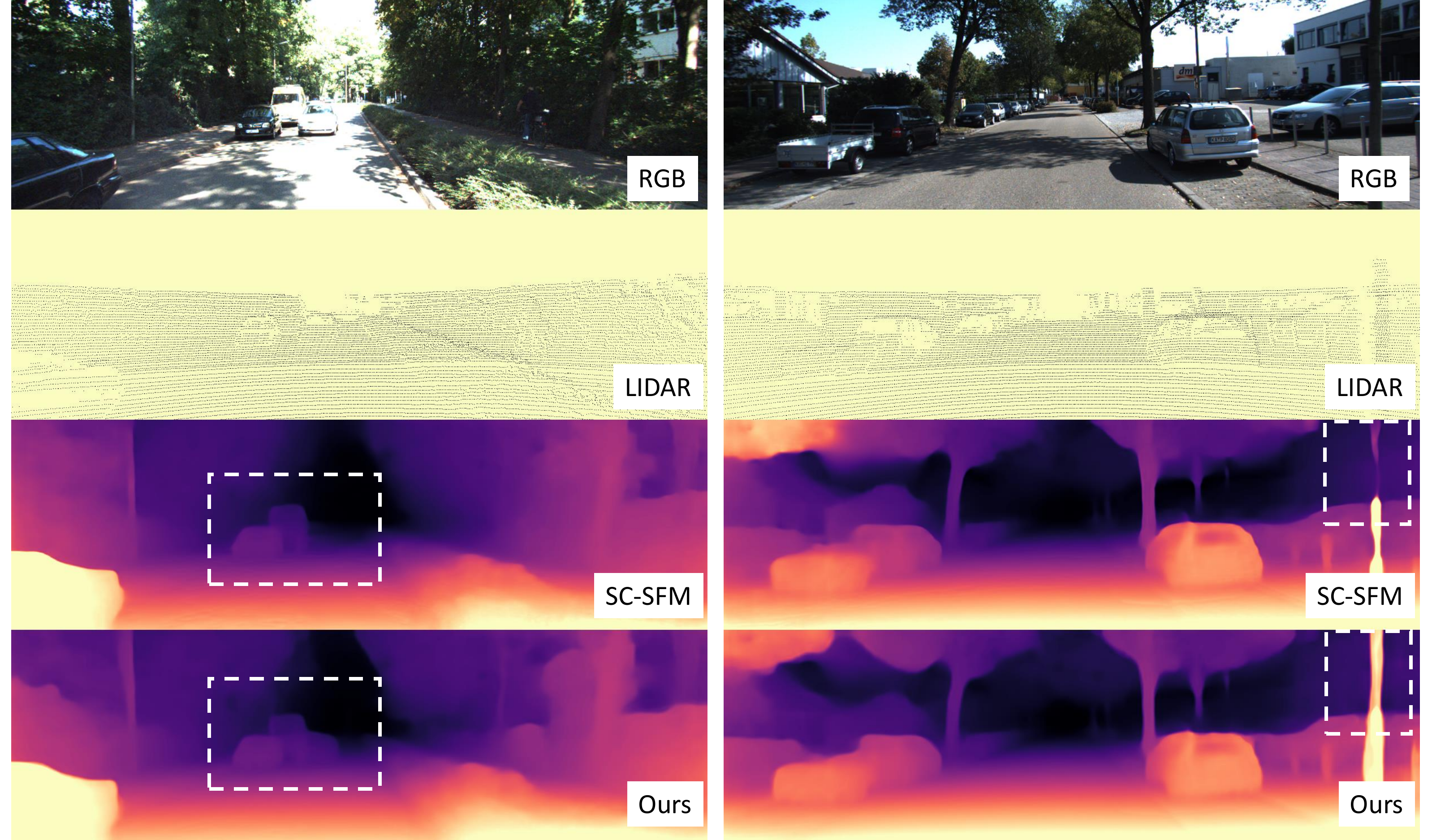}
\caption{Samples of monocular depth estimation results on the KITTI dataset for qualitative comparison of the unsupervised  methods. Our DPVO captures details in challenging scenes that contain occlusions and uneven road lines. Some examples of these important differences are highlighted with dashed boxes.}
\label{fig:res_depth_kitti}
\end{figure}

\begin{table*}[t]
    \centering
    \setlength{\tabcolsep}{10.0pt}
    \begin{tabular}{l |c| c c c c | c c c}
    \hline
     & & \multicolumn{4}{c|}{Error $\downarrow$} & \multicolumn{3}{c}{Accuracy $\uparrow$}  \\
     \cline{3-9}
     Methods & Dataset & AbsRel & SqRel & RMS & RMSlog & $<1.25$ & $<1.25^2$ & $<1.25^3$ \\
     \hline
     Zhou et al.~\cite{zhou2017unsupervised} & K  & 0.208 & 1.768 & 6.856 & 0.283 & 0.678 & 0.885 & 0.957 \\
     Mahjourian et al.~\cite{mahjourian2018unsupervised} & K  & 0.163 & 1.240 & 6.220 & 0.250 & 0.762 & 0.916 & 0.968 \\
     Geonet~\cite{yin2018geonet}  & K  & 0.155 & 1.296 & 5.857 & 0.233 & 0.793 & 0.931 & 0.973\\
     DF-Net~\cite{zou2018df}  & K  & 0.150 & 1.124 & 5.507 & 0.223 & 0.806 & 0.933 & 0.973\\
     CC~\cite{ranjan2019competitive}  & K  & 0.140 & \textbf{1.070} & 5.326 & 0.217 & 0.826 & 0.941 & \textbf{0.975}\\
     GANVO \cite{almalioglu2019ganvo} & K & 0.150 & 1.141 & 5.448 & 0.216 & 0.808 & 0.939 & \textbf{0.975}\\
     SC-SFM \cite{bian2019unsupervised} & K  & 0.137 & 1.089 & 5.439 & 0.217 & 0.830 & \textbf{0.942} & \textbf{0.975}\\
     Ours & K  & \textbf{0.127} & 1.077 & \textbf{5.312} & \textbf{0.214} & \textbf{0.835} & 0.941 & \textbf{0.975}\\
     \hline
     Zhou et al.~\cite{zhou2017unsupervised} & CS+K  & 0.198 & 1.836 & 6.565 & 0.275 & 0.718 & 0.901 & 0.960 \\
     Mahjourian et al.~\cite{mahjourian2018unsupervised} & CS+K  & 0.159 & 1.231 & 5.912 & 0.243 & 0.784 & 0.923 & 0.970 \\
     Geonet~\cite{yin2018geonet}  & CS+K  & 0.153 & 1.328 & 5.737 & 0.232 & 0.802 & 0.934 & 0.972 \\
     DF-Net~\cite{zou2018df}  & CS+K  & 0.146 & 1.182 & 5.215 & 0.213 & 0.818 & 0.943 & \textbf{0.978} \\
     CC~\cite{ranjan2019competitive}  & CS+K  & 0.139 & \textbf{1.032} & 5.199 & 0.213 & 0.827 & 0.943 & 0.977 \\
     GANVO \cite{almalioglu2019ganvo} & CS+K & 0.138 & 1.155 & 4.412 & 0.232 & 0.820 & 0.939 & 0.976\\
     SC-SFM \cite{bian2019unsupervised} & CS+K  & 0.128 & 1.047 & 5.234 & \textbf{0.208} & 0.846 & 0.947& 0.976 \\
     Ours & CS+K  & \textbf{0.122} & 1.039 & \textbf{5.184} & \textbf{0.208} & \textbf{0.851} & \textbf{0.948} & 0.976 \\
     \hline
     CC~\cite{ranjan2019competitive}  & SubT  & 0.214 & 1.486 & 6.280 & 0.284 & 0.713 & 0.912 & 0.952 \\
     GANVO \cite{almalioglu2019ganvo} & SubT & 0.190 & 1.391 & 5.899 & 0.266 & 0.746 & 0.920 & 0.962\\
     SC-SFM \cite{bian2019unsupervised} & SubT  & 0.175 & \textbf{1.309} & 5.772 & 0.260 & 0.765 & 0.925 & 0.964 \\
     Ours & SubT  & \textbf{0.149} & 1.338 & \textbf{5.484} & \textbf{0.229} & \textbf{0.792} & \textbf{0.935} & 0.969 \\
    \hline
    \end{tabular}
     \caption{Monocular single-view depth estimation results, testing on the odometry split of KITTI dataset~\cite{geiger2013vision}. The methods trained on KITTI raw~\cite{geiger2013vision} and the DARPA subterranean datasets are denoted by K and SubT, respectively. Models with pre-training on CityScapes~\cite{cordts2016cityscapes} are denoted by CS+K. The best performance in each block is highlighted with bold font.}
        \label{tab:depth}
\end{table*}

\begin{figure*}[t]
\centering
\includegraphics[width=\textwidth]{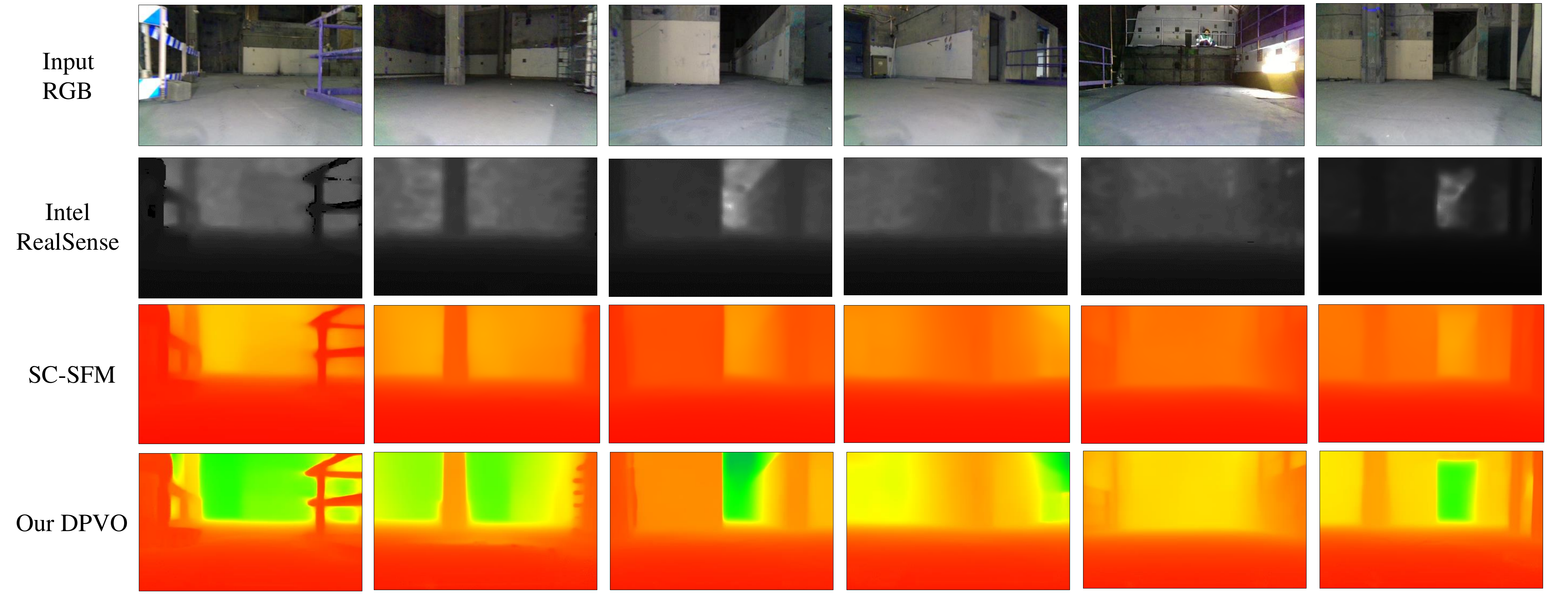}
\caption{Qualitative comparison of two unsupervised monocular depth estimation methods on the challenging DARPA subterranean dataset. The stereo depth output of the Intel RealSense camera (D435i model) is shown for visual comparison purposes. Our DPVO captures details in challenging scenes containing low textured areas, poorly illuminated regions, and with strong occlusions, preserving accurate and detailed depth map predictions both in close and distant regions.}
\label{fig:res_depth_subt}
\end{figure*}

\subsection{Single-view depth evaluation}

Our proposed approach produces (and in most cases improves) state-of-the-art results on single view depth predictions, as shown in Table \ref{tab:depth}.
Here, the depth is evaluated on the Eigen et al.~\cite{eigen2014depth} split of the raw KITTI dataset~\cite{geiger2013vision} following the previous works~\cite{eigen2014depth, liu2016learning, mahjourian2018unsupervised, yin2018geonet}.
As shown in Table~\ref{tab:depth}, our method outperforms the other competitors~\cite{almalioglu2019ganvo, bian2019unsupervised, ranjan2019competitive} on several benchmarks. 
Previous works in the literature~\cite{almalioglu2019ganvo, bian2019unsupervised, gordon2019depth} proved that transfer learning from Cityscapes dataset to KITTI is beneficial and leads to more accurate depth estimation; thus we include CS+K benchmark in this work to compare cross-dataset generalizability of our DPVO. 
DPVO significantly improves the performance on depth estimation benchmarks using Cityscapes in the training (see CS+K in Table~\ref{tab:depth}).

Figure \ref{fig:res_depth_kitti} shows examples of depth map results predicted by our DPVO and SC-SFM methods along with the RGB input and ground-truth. 
We highlight the notable differences with SC-SFM, which fails to capture distant objects in the scene.
Furthermore, Fig.~\ref{fig:res_depth_kitti} also shows that the depth maps predicted by the proposed DPVO capture the small objects in the scene, whereas the other methods tend to ignore them. 
Most importantly, as shown in the bottom rows of Table \ref{tab:depth} (quantitatively) and in Fig.~\ref{fig:res_depth_subt} (qualitatively), our unsupervised approach significantly outperforms state-of-the-art methods in challenging scenarios.
The proposed DPVO also accurately predicts the depth values of the objects in low-textured areas caused by the perceptual degradation in a scene.
A simple loss function on the depth map without handling occlusions leads to averaging all likely locations of details, whereas the depth enhancement modules in feature space with a natural depth prior and geometric loss constraints make the proposed DPVO more sensitive to the likely positions of the details in the scene.

\section{CONCLUSIONS}
\label{sec:conclusion}
In this study, we proposed an unsupervised deep learning method for pose and depth map estimation using monocular image sequences. 
This work addresses critical challenges for unsupervised learning of depth and visual odometry through geometric occlusion-aware and scale-aware loss functions as well as depth enhancement modules.
The proposed method outperforms all the competing unsupervised and traditional baselines in terms of pose estimation and depth map reconstruction by a significant margin in challenging environments.
As a path forward, we plan to explicitly address optical flow in order to improve the performance in such perception-challenging environments.


\bibliographystyle{IEEEtran}
\balance

\bibliography{bibfile}

\end{document}